\documentclass{article}

\usepackage{iclr2026_conference,times}

\PassOptionsToPackage{numbers, compress}{natbib}

\usepackage[utf8]{inputenc} 
\usepackage[T1]{fontenc}    
\usepackage{hyperref}       
\usepackage{url}            
\usepackage{booktabs}       
\usepackage{amsfonts}       
\usepackage{nicefrac}       
\usepackage{microtype}      
\usepackage{xcolor}         
\usepackage{amsmath}
\usepackage{subcaption}
\usepackage{caption}
\usepackage{listings}
\usepackage{xspace}
\usepackage{cleveref}

\usepackage{graphicx}
\usepackage{duckuments}
\usepackage[subtle]{savetrees}

\newcommand\sL{\ensuremath{\mathcal{L}}}
\newcommand\sM{\ensuremath{\mathcal{M}}}

\newcommand\sX{\ensuremath{\mathcal{X}}}
\newcommand\sY{\ensuremath{\mathcal{Y}}}
\newcommand\sZ{\ensuremath{\mathcal{Z}}}

\newcommand\BR{\ensuremath{\mathbb{R}}}

\newcommand\eqnl[2]{\begin{align} \label{eqn:#1} #2 \end{align}} 

\ifthenelse{\isundefined{\definition}}{\newtheorem{definition}{Definition}}{}
\ifthenelse{\isundefined{\assumption}}{\newtheorem{assumption}{Assumption}}{}
\ifthenelse{\isundefined{\hypothesis}}{\newtheorem{hypothesis}{Hypothesis}}{}
\ifthenelse{\isundefined{\proposition}}{\newtheorem{proposition}{Proposition}}{}
\ifthenelse{\isundefined{\theorem}}{\newtheorem{theorem}{Theorem}}{}
\ifthenelse{\isundefined{\lemma}}{\newtheorem{lemma}{Lemma}}{}
\ifthenelse{\isundefined{\corollary}}{\newtheorem{corollary}{Corollary}}{}
\ifthenelse{\isundefined{\alg}}{\newtheorem{alg}{Algorithm}}{}
\ifthenelse{\isundefined{\example}}{\newtheorem{example}{Example}}{}
\newcommand{\E}{\ensuremath{\mathbb{E}}} 

\newcommand\map[2]{#1\rightarrow#2}
\newcommand\pstar{p^*}
\newcommand{\expvalxy}{\E_{x\sim\sX,y\sim\pstar(\cdot \mid x)}}

\newcommand{\expvalxmask}{\E_{x\sim\sX,m \sim \sM}}
\newcommand{\expvalxx}{\E_{x_i,x_{i+1}\sim\sX}}
\newcommand{\lossvicreg}{\ell_{\text{VICReg}}}
\newcommand{\lossft}{\ell_{\text{ft}}}
\newcommand{\norm}[1]{\left\lVert #1 \right\rVert}

\def\shownotes{1}  
\ifnum\shownotes=1
\newcommand{\authnote}[2]{[#1: #2]}
\else
\newcommand{\authnote}[2]{}
\fi

\title{Representation Learning for Spatiotemporal Physical Systems}

\author{
\hfil \begin{minipage}{\textwidth}
    \vspace{3ex}
    \centering
    {\bf Helen Qu$^{1}$\thanks{Contact: \texttt{hqu@flatironinstitute.org}} \quad
    Rudy Morel$^{1}$ \quad
    Michael McCabe$^{1,4}$ \quad
    Alberto Bietti$^{1}$ \quad
    François Lanusse$^{2}$} \\
    \vspace{1ex}
    {\bf Shirley Ho$^{1,3,4}$ \quad
    Yann LeCun$^{3}$} \\
    \vspace{2ex}
    {\bf The Polymathic AI Collaboration} \\
    \vspace{2ex}
    \textmd{
        \begin{tabular}{l}
        $^{1}$Flatiron Institute \\
        $^{2}$Université Paris-Saclay, Université Paris Cité, CEA, CNRS, AIM \\
        $^{3}$New York University\\
        $^{4}$Princeton University
        \end{tabular}
    }
\end{minipage}
}

\iclrfinalcopy
\begin{document}

\maketitle

\begin{abstract}

Machine learning approaches to spatiotemporal physical systems have primarily focused on next-frame prediction, with the goal of learning an accurate emulator for the system's evolution in time.
However, these emulators are computationally expensive to train and are subject to performance pitfalls, such as compounding errors during autoregressive rollout.
In this work, we take a different perspective and look at scientific tasks further downstream of predicting the next frame, such as estimation of a system's governing physical parameters.
Accuracy on these tasks offers a uniquely quantifiable glimpse into the physical relevance of the representations of these models.
We evaluate the effectiveness of general-purpose self-supervised methods in learning physics-grounded representations that are useful for downstream scientific tasks.
Surprisingly, we find that not all methods designed for physical modeling outperform generic self-supervised learning methods on these tasks, and methods that learn in the latent space (e.g., joint embedding predictive architectures, or JEPAs) outperform those optimizing pixel-level prediction objectives.\footnote{Code: \texttt{https://github.com/helenqu/physical-representation-learning}.}

\end{abstract}

\section{Introduction}

Understanding and forecasting of physical systems is a challenging problem with applications ranging from biological development to astrophysical phenomena \citep{maddu2024learning,morelpredicting}.
Many applications of machine learning in this space explore autoregressive surrogate modeling, which aims to learn frame-by-frame, pixel-by-pixel emulators of computationally expensive numerical simulations \citep[e.g.,][]{mccabe2023multiple,herde2024poseidon,mccabe2025walrus}.
However, these full-field prediction models are computationally expensive to train and may not be best suited for higher-level downstream tasks of scientific interest, such as parameter estimation or qualitative prediction (e.g., whether the system remains laminar or becomes turbulent).
Relatively little attention has been paid to understanding which learning paradigms optimally learn and preserve physically meaningful information.

\begin{figure}
    \centering
    \includegraphics[width=0.7\linewidth,trim=0.2cm 0 0 0cm,clip]{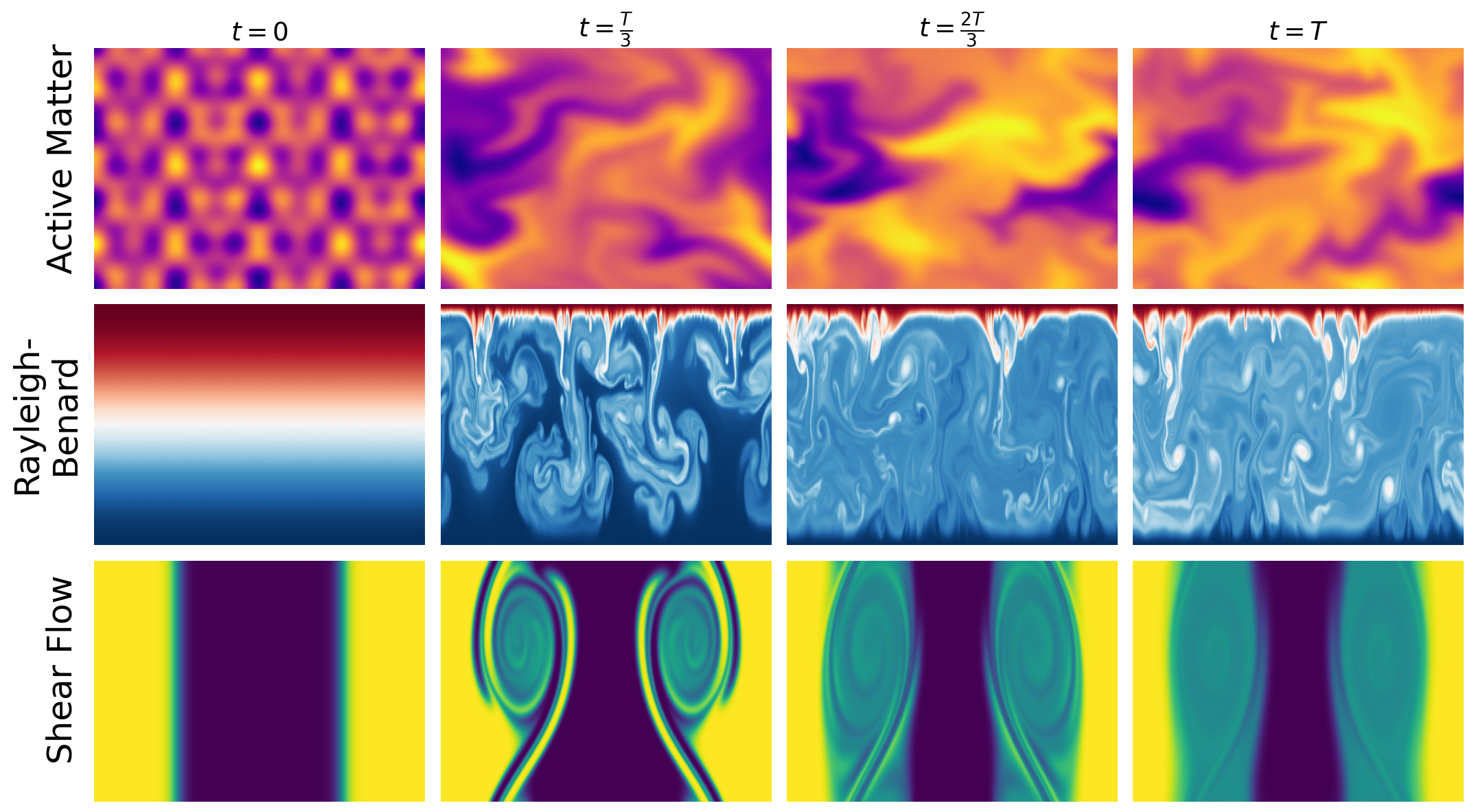}
    \caption{Example trajectories from the physical systems in our evaluation.}
    \label{fig:systems}
\end{figure}

In this work, we investigate the efficacy of self-supervised learning paradigms for tackling scientifically meaningful tasks in spatiotemporal physical systems.
We compare traditional masked autoencoding and joint embedding predictive architectures \citep[JEPAs,][]{lecun2022path,assran2023self,bardes2024revisiting,assran2025v} with methods developed for physical modeling and inference on three representative physical systems (active matter, shear flow, and Rayleigh-Bénard convection; see Figure~\ref{fig:systems}).
Unlike traditional self-supervised methods, JEPAs are trained to predict in the model's learned latent space rather than the granular, low-level space of pixel values.
We probe the models’ understanding of these systems through physical parameter estimation, a quantifiable proxy for physical information, and demonstrate that embeddings learned by latent prediction models consistently outperform pixel-based self-supervised methods.

\section{Setup}
\label{gen_inst}

We introduce the self-supervised representation learning methods, physical modeling baselines, and fine-tuning procedure used in this work.

\subsection{Representation Learning Frameworks}
\paragraph{Joint Embedding Predictive Architectures.} 
We introduce a latent feature prediction model based on joint embedding predictive architectures (JEPA) optimized for dynamics.
JEPAs minimize error in the space of representations, while pixel reconstruction models learn low-level detail in the visual input that may not be helpful for understanding the simple underlying dynamics.
Our JEPA formulation posits that for a sample with $T$ temporal steps $x_{0:T}$, aligning the representation of $x_{t:t+k}$ with the representation of $x_{t+k:t+2k}$ (i.e. predicting the representation of the next sequence of $k$ frames from a context sequence of $k$ frames) will produce flexible embeddings that carry high-level information about the visual content.
Formally, we aim to learn an encoder $f: \map{\sX}{\sZ}$ and predictor $g: \map{\sZ}{\sZ}$ that minimizes $\sL(f,g) = \expvalxx [ \lossvicreg (g(f(x_i)), f(x_{i+1})]$, where $\lossvicreg$ is defined following \citet{bardes2021vicreg} to prevent mode collapse:
\eqnl{vicreg}{\lossvicreg (z_i, z_{i+1}) = \lambda s(z_i, z_{i+1}) + \mu[v(z_i) + v(z_{i+1})] + \nu[c(z_i) + c(z_{i+1})]}
Here, $s(z_i,z_{i+1})=\frac{1}{n}\sum_{i} \norm{z_{i+1} - z_i}^2_2$ is the invariance criterion, $v(z)=\frac{1}{d}\sum_{j=1}^d\text{max}(0, 1-\sqrt{\text{var}(z)+\epsilon})$ is the variance regularization term, and $c(z)=\frac{1}{d}\sum_{i \neq j} C_{ij}(z)^2$ is the covariance regularization term, where $C$ is the covariance matrix of $z$. $\lambda, \mu, \nu$ are hyperparameters controlling the importance of each term in the loss, and $d$ is the batch size.

\paragraph{Masked autoencoding.} Autoencoders aim to learn an encoder-decoder pair $f: \map{\sX}{\sZ}, g: \map{\sZ}{\sX}$, where $\sZ$ is the abstract representation space of the model, that minimizes pixel-level reconstruction error.
We implement masked autoencoding in this work, where reconstruction error is computed over masked regions $x(m), m \in \{0,1\}^N$: $\sL(f,g) = \expvalxmask [\hat{x}(m) - x(m))^2]$, where $\hat{x} = g(f(x))$.
In practice, masked autoencoders have been extended to spatiotemporal data by enforcing temporal tube masking, i.e. all frames $x_{0:T}$ use the same spatial mask $m$.
Autoencoding is a common paradigm for feature learning from large datasets.

\subsection{Physical Modeling Baselines}
Our baselines center their approaches on learning physical priors, either through data (in the style of autoregressive foundation models) or through architectural/training inductive biases (neural operator-style approaches).

\paragraph{Autoregressive foundation models.}
Foundation models for physics, often implemented as pixel-level autoregressive models for spatiotemporal systems, learn to predict the pixel values of the frame at the next timestep $x_{t+1}$ given a context window of the previous $n$ frames, $x_{t-n:t}$. These are often called ``surrogate models'' due to their potential to replace the computationally expensive procedure of numerically solving for the next step of the spatiotemporal system (e.g., through PDE numerical solvers).

\paragraph{In-context operator learning.} 
In-context operator learning models combine in-context transformers with the inductive bias of neural operators. Rather than learning a single operator across all trajectories, that are described by multiple physics, this approach infers a trajectory-specific operator network $f_\theta$, that is, an evolution rule, from a short context window and evolves it forward in time using explicit integration.

\subsection{Fine-tuning} Fine-tuning learns a prediction head $h: \map{\sZ}{\sY}$ on embeddings from the pretrained model to optimize $\sL(h) = \expvalxy [\lossft (h(f(x)), y)]$, where $\lossft: \map{\sY \times \sY}{\BR}$ is implemented as squared error loss for regression.

\section{Evaluating Representations with Physical Systems}
\label{sec:data}
We evaluate on the task of physical parameter prediction: the minimum prediction error achievable on physical parameters underlying each system. Intuitively, these parameters govern the time evolution of these systems, so low inference error indicates better understanding of the underlying physical process. We show some examples of how evolution depends on the values of these parameters in Figure~\ref{fig:diff-params}. We evaluate across three PDE-governed spatiotemporal systems drawn from The Well \citep{ohana2025welllargescalecollectiondiverse}.

\paragraph{Active matter.}
Active matter systems are a collection of agents that convert chemical energy into mechanical work, causing emergent, system-wide patterns and collective dynamics. 
This dataset models the dynamics of $N$ rodlike active particles immersed in a Stokes fluid, which is modeled by kinetic theory.
The system parameters of interest are $\alpha$, the active dipole strength, and $\zeta$, the strength of particle alignment through steric interactions.

\paragraph{Rayleigh-Bénard convection.}
This system describes the behavior of a horizontal fluid layer heated from below and cooled from above, forming Bénard convective cells due to the temperature gradient. 
The system parameters of interest parameterize properties of the fluid layer: the Rayleigh number $\nu$, the ratio of buoyancy forces to viscous forces, and the Prandtl number $\kappa$, the ratio of momentum diffusivity to thermal diffusivity.

\paragraph{Shear flow.}
Shear flow describes the boundary between layers of fluid (modeled by incompressible Navier-Stokes) moving parallel to each other at different velocities, potentially leading to vortex/eddy formation and turbulence.
The system parameters of interest are the Reynolds number, the ratio between inertial and viscous forces in the fluid, and the Schmidt number, the ratio of momentum diffusivity to mass diffusivity, of the fluids.

\section{Experiments}
We use our physical dynamics testbed to compare the representations learned by our latent dynamics JEPA model with those of a masked autoencoder trained with masked pixel prediction objective, which we implement in practice as a VideoMAE ViT-tiny/16 \citep{Tong2022_vmae}. We additionally compare to two baselines: the operator meta-learning framework DISCO \citep{morel2025discolearningdiscoverevolution} and the autoregressive model MPP \citep{mccabe2024multiple}.

\paragraph{Pretraining procedure.} We implement the JEPA encoder as a downsampling CNN following ConvNeXt \citep{liu2022convnet}, while the predictor is a CNN with an inverse bottleneck in the channel dimension. We pretrain VideoMAE from scratch following \citet{Tong2022_vmae}. We pretrain separate JEPA and VideoMAE models on each physics dataset individually to encourage representations optimized for each system's unique dynamics. MPP is intended as a foundation model approach, so we use the published pretrained weights for their AViT-tiny model. Finally, we use a DISCO model pretrained on The Well following \citet{morel2025discolearningdiscoverevolution}. Further implementation details are provided in Appendix~\ref{appx:implementation}.

\paragraph{Fine-tuning procedure.} For VideoMAE, JEPA, and DISCO models, we follow the procedure outlined in \citet{bardes2024revisiting} to fine-tune attentive probes for 100 epochs on top of the frozen encoders. We keep the encoder weights frozen to evaluate the physically meaningful information each model was able to learn without explicit supervision. We perform end-to-end fine-tuning of MPP following \citet{mccabe2023multiple}, since unlike the other models, the MPP pretraining did not include the active matter, shear flow, and Rayleigh-Bénard datasets. As described in Section~\ref{sec:data}, we report the averaged MSE on parameters $\alpha$ and $\zeta$ for active matter, Reynolds and Schmidt numbers for shear flow, and Rayleigh and Prandtl numbers for Rayleigh-Bénard.

\paragraph{JEPAs outperform MAE in parameter estimation.} We show results for parameter estimation in Table~\ref{tbl:params} in terms of mean-squared-error loss (MSE, lower is better). JEPA improves substantially on VideoMAE results across the board. We find a 51\% relative improvement on active matter ($0.16 \rightarrow 0.08$), 43\% on shear flow ($0.67 \rightarrow 0.38$), and 28\% ($0.18 \rightarrow 0.13$) on Rayleigh-Bénard. 

\paragraph{Comparison with methods for physical modeling.} We are interested in comparing our generic representation learning methods against the goalposts of physics models like DISCO and MPP. We find that, there is large variance in the efficacy of using these methods with representation learning recipes. Fine-tuning from frozen DISCO latent representations yield excellent parameter prediction results, while MPP struggles despite end-to-end fine-tuning. This is consistent with \citet[][Table 9]{mccabe2023multiple}, and also consistent with results from the language modeling community demonstrating that autoregressive modeling approaches generally underperform encoder-only approaches on non-generative tasks \citep[e.g.,][]{devlin2018bert,raffel2020exploring}. It is also interesting to note the variations between the methods for certain systems/tasks. For example, DISCO and JEPA perform very similarly (MSE of 0.057 and 0.079, respectively) on the active matter dataset, while differing by an order of magnitude (0.01 and 0.13) on Rayleigh-Bénard. However, VideoMAE closes the gap with JEPA most effectively on Rayleigh-Bénard (0.13 vs. 0.18). Finally, we note that DISCO and JEPA are both latent prediction models, while MPP and VideoMAE are pixel-level prediction models, and our results show that DISCO and JEPA are the two best performing models for their respective classes.

\begin{table}[]
\centering
\caption{Physical parameter prediction MSE of the self-supervised (top 2 rows) and physical modeling (bottom 2 rows) methods after fine-tuning. JEPA outperforms VideoMAE on physical parameter estimation, and JEPA prediction error approaches that of DISCO, the best physical modeling method we tested.}
\label{tbl:params}
\begin{tabular}{@{}lccc@{}}
\toprule
                   & \multicolumn{3}{c}{MSE ($\downarrow$)} \\
                      & active matter & shear flow & Rayleigh-Bénard convection \\ \midrule
JEPA                  & \textbf{0.079  }       & \textbf{0.38}       & \textbf{0.13}                       \\
VideoMAE              & 0.160         & 0.67       & 0.18                       \\  \midrule
DISCO                 & \textbf{0.057 }        & \textbf{0.13}       & \textbf{0.01}                       \\
MPP (full finetuning) & 0.230         & 0.59       & 0.08         \\
\bottomrule
\end{tabular}
\end{table}

\begin{table}[]
\centering
\caption{Physical parameter prediction MSE with increasing fine-tuning dataset size, tested on the shear flow parameter prediction task. JEPA exhibits better data scaling behavior compared to VideoMAE.}
\label{tbl:params-ft-efficiency}
\begin{tabular}{@{}lccc@{}}
\toprule
& \multicolumn{3}{c}{fine-tuning dataset fraction} \\
         & 10\%          & 50\%         & 100\%         \\ \midrule
JEPA     & \textbf{0.57} & \textbf{0.40} & \textbf{0.38} \\
VideoMAE & 0.98          & 0.75         & 0.67      \\
\bottomrule
\end{tabular}
\end{table}

\paragraph{JEPA exhibits desirable scaling behavior for fine-tuning data.} We compare the sample efficiency of JEPA and VideoMAE at fine-tuning time in Table~\ref{tbl:params-ft-efficiency}. We show using the shear flow parameter estimation task that with just 50\% of the available fine-tuning data (16k examples), JEPA attains an MSE loss of 0.4 (95\% of the best performance, 0.38). VideoMAE, however, exhibits a larger drop between 50\% and 100\% (0.75, which is 89\% of the best performance, 0.67). Moreover, even with 10\% of the fine-tuning data, JEPA outperforms VideoMAE's best performance with 100\% of the data (0.57 compared to 0.67).

\section{Conclusion}
We investigated the ability of general-purpose self-supervised methods to learn physically meaningful information in spatiotemporal systems by evaluating their ability to recover governing parameters across three PDE benchmarks. Our results show that latent prediction objectives consistently produce representations that are more physically informative and more sample-efficient to fine-tune compared to pixel-level reconstruction and autoregressive models. These findings highlight the value of disentangling the physical relevance of learned representations from generative fidelity and suggest that alternatives to autoregressive surrogate modeling, such as latent-space predictive learning, could be a promising foundation for scientific machine learning.

\subsubsection*{Acknowledgments}
We thank the Scientific Computing Core at the Flatiron Institute, a division of the Simons
Foundation, for providing computational resources and support. We also acknowledge Jeremy Cohen, Tanya Marwah, Sebastian Wagner-Carena, the Polymathic AI team, and our anonymous reviewers for helpful discussions and comments. Finally, Polymathic AI gratefully acknowledges funding from the Simons Foundation and Schmidt Sciences.

\bibliography{walrus,extra,references}
\bibliographystyle{iclr2026_conference}

\clearpage
\appendix

\begin{figure}
    \centering
    \includegraphics[width=1\linewidth]{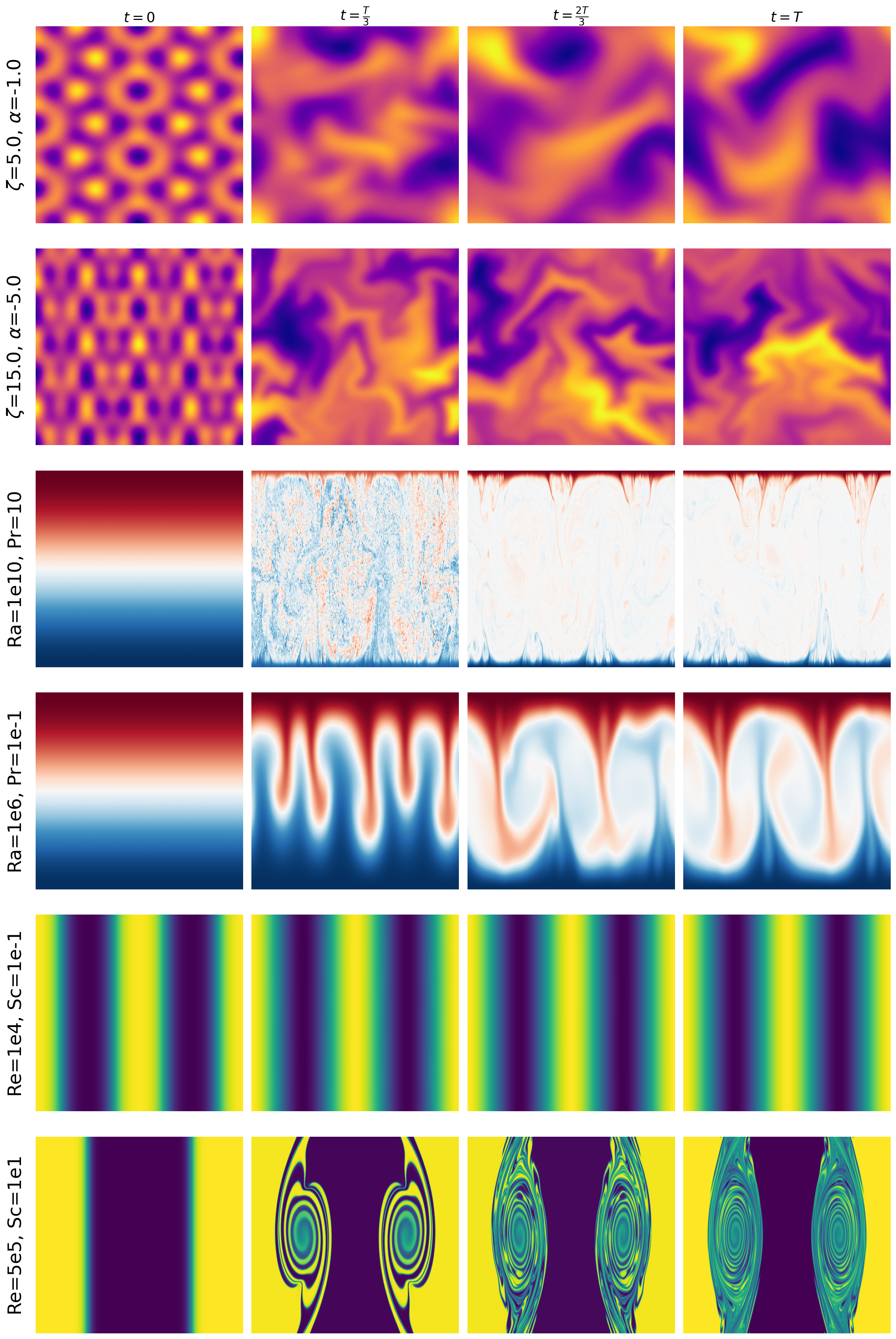}
    \caption{Different values for the physical parameters, e.g., Rayleigh (Ra) and Prandtl (Pr) numbers for Rayleigh-Bénard, lead each system to evolve very differently.}
    \label{fig:diff-params}
\end{figure}

\section{Related Work}
\paragraph{Machine learning for physical systems.} 
Predicting physical dynamics has motivated a wide range of approaches \citep{sirignano2018dgm, yu2018deep, han2018solving, bar2019unsupervised, zang2020weak} with varying amounts of assumed knowledge about the governing equations. More recently, there has been growing interest in developing foundation models for dynamical systems, training on a large corpus of physical data and adapting the network to various downstream tasks through fine-tuning \citep[e.g.,][]{mccabe2023multiple,herde2024poseidon,nguyen2025physixfoundationmodelphysics,sun2025foundationmodelpartialdifferential}. However, most of these approaches focus on modeling the dynamics while very little work has been done with an eye towards concrete downstream scientific tasks. The most relevant work to ours is \cite{mialon2023self}, which introduces a Lie symmetries-based augmentation procedure for self-supervised representation learning on physical systems. In our work, we explore the efficacy of various self-supervised as well as physical modeling objectives through the lens of representation learning.

\paragraph{Self-supervised learning.}
Self-supervised learning objectives allow models to learn generally useful representations from data without the need for task-specific human annotations \citep{balestriero2023cookbook}. This is done through the introduction of a pretext task, e.g., next token prediction in modern language models \citep[e.g.,][]{openai2023gpt4,team2023gemini,grattafiori2024llama3}, masked autoencoding \citep{he2021masked,devlin2018bert}, contrastive methods \citep{chen2020simclr,chen2020improved,caron2020swav}, self-distillation \citep{caron2021emerging}, and joint embedding architectures \citep{assran2023self,bardes2024revisiting,assran2025v}.
However, these are primarily evaluated with human-centric ``natural'' data (e.g., ImageNet \citep{deng2009imagenet}).
This work investigates general-purpose self-supervised learning frameworks in the context of scientific data, specifically in a setting for which ground-truth governing equations and parameters are known.
This unique testbed gives a new perspective on the representations learned by these methods.

\section{Implementation Details}
\label{appx:implementation}
All encoders are given input sequences of $l \times w \times d \times t$, where $l,w$ are the spatial dimensions of the image, $d$ is the number of physical fields (e.g., buoyancy, pressure, etc.) and $t=16$ context frames.

The JEPA encoder and predictor architectures are 3D convolutional neural networks. The final encoder output is $l/16 \times w/16 \times 128$. We use the ``small'' VideoMAE ViT architecture with patch size 16 provided in their codebase, with encoder output $l/16 \times w/16 \times t/2 \times 384$. We use the output of DISCO's hypernetwork as embeddings, which are $1 \times 384$, and MPP's embeddings are $l/16 \times w/16 \times 192$.

JEPA and VideoMAE are pretrained for 6 epochs, and all models are finetuned for 100 epochs. We use the AdamW optimizer with a cosine learning rate schedule for all training/fine-tuning.

Empirically, we find good performance by choosing hyperparameters $\lambda=2, \mu=40, \nu=2$ for the VICReg-style loss function used to pretrain the JEPA models.

\end{document}